\documentclass[letterpaper, 10 pt, conference]{ieeeconf}  

\IEEEoverridecommandlockouts                              

\usepackage{pgfplots}
\pgfplotsset{compat=1.18}
\definecolor{fraunhoferOrange}{RGB}{245,130,32} 
\definecolor{fraunhoferGreen}{RGB}{23, 156, 125} 
\definecolor{fraunhoferBlue}{RGB}{0, 91, 127} 
\definecolor{fraunhoferGrey}{RGB}{166,187,200} 
\definecolor{fraunhoferTurquoise}{RGB}{0,133,152} 
\definecolor{fraunhoferSkyBlue}{RGB}{57,193,205} 
\definecolor{fraunhoferMint}{RGB}{178,210,53} 
\usepackage{xcolor}

\usepackage{tikz}
\usetikzlibrary{shapes, arrows}\usepackage{float}
\usepackage{caption}
\usepackage{subcaption}
\usepackage{todonotes}
\usepackage{amsmath,amssymb,amsfonts}
\usepackage{algorithmic}
\usepackage{graphicx}
\usepackage{algorithm,algorithmic}
\usepackage{hyperref}
\usepackage{array}
\usepackage{multirow}
\hypersetup{hidelinks=true}
\usepackage{textcomp}
\usepackage{booktabs}
\usepackage{mathtools}
\usepackage{comment}
\usepackage{cite} 
\usepackage{amsmath}
\usepackage{cite} 
\usepackage{hyperref}
\PassOptionsToPackage{pdftex}{graphicx}
\usepackage{graphicx}
\usepackage{verbatim}
\graphicspath{{./pdf/}{./jpeg/}{./img/}}

\title{\LARGE \bf
Improving AI-Based Canine Heart Disease Diagnosis with Expert-Consensus Auscultation Labeling}

\author{Pinar Bisgin$^{1,2}$, Tom Strube$^{1}$, Niklas Tschorn$^{1,6}$, Michael Pantförder$^{1,6}$, Maximilian Fecke$^{1,2}$,\\ 
Ingrid Ljungvall$^{3}$, Jens Häggström$^{3}$, Gerhard Wess$^{4}$, Christoph Schummer$^{5}$, Sven Meister$^{6}$, Falk M. Howar$^{2}$
\newline
\thanks{$^{1}$Fraunhofer Institute for Software and Systems Engineering ISST, Dortmund, Germany. Pinar.Bisgin@isst.fraunhofer.de
}%
\thanks{$^{2}$Department of Computer Science, TU Dortmund University, Dortmund, Germany. 
}%
\thanks{$^{3}$Department of Clinical Sciences, Swedish University of Agricultural Sciences, Uppsala, Sweden. 
}%
\thanks{$^{4}$Clinic for Small Animal Medicine, Ludwig Maximilian University of Munich, Germany. 
}%
\thanks{$^{5}$Boehringer Ingelheim Vetmedica GmbH, Ingelheim am Rhein, Germany. 
}%
\thanks{$^{6}$Department of Health Informatics, School of Medicine, Witten/Herdecke University, Witten, Germany. 
}%
}

\begin{document}

\maketitle
\thispagestyle{empty}
\pagestyle{empty}

\begin{abstract}
Noisy labels pose significant challenges for AI model training in veterinary medicine. This study examines expert assessment ambiguity in canine auscultation data, highlights the negative impact of label noise on classification performance, and introduces methods for label noise reduction.
To evaluate whether label noise can be minimized by incorporating multiple expert opinions, a dataset of $140$ heart sound recordings (HSR) was annotated regarding the intensity of holosystolic heart murmurs caused by Myxomatous Mitral Valve Disease (MMVD).
The expert opinions facilitated the selection of $70$ high-quality HSR, resulting in a noise-reduced dataset. 
By leveraging individual heart cycles, the training data was expanded and classification robustness was enhanced.
The investigation encompassed training and evaluating three classification algorithms: AdaBoost, XGBoost, and Random Forest. 
All of them showed significant improvements in classification accuracy due to the applied label noise reduction, most notably XGBoost.
Specifically, for the detection of mild heart murmurs, sensitivity increased from $37.71\%$ to $90.98\%$ and specificity from $76.70\%$ to $93.69\%$. 
For the moderate category, sensitivity rose from $30.23\%$ to $55.81\%$ and specificity from $64.56\%$ to $97.19\%$. 
In the loud/thrilling category, sensitivity and specificity increased from $58.28\%$ to $95.09\%$ and from $84.84\%$ to $89.69\%$, respectively.
These results highlight the importance of minimizing label noise to improve classification algorithms for the detection of canine heart murmurs.
\end{abstract}


\section{INTRODUCTION}
\label{sec:introduction}
In veterinary medicine, machine learning algorithms often rely on objective data, such as physiological measurements, with subjective assessments as labels. Noise in both data and labels can impair prediction performance, leading to reduced reliability \cite{frenay2013classification}. 
In the context of auscultation, a central method for disease diagnosis, data collection is particularly susceptible to noise. 
Factors such as fur, panting, and morphological variations can obscure the analyzed physiological sounds, while inconsistent diagnostic annotations may obscure the corresponding labels. 
Both types of noise must be considered in AI-powered decision support systems to prevent misdiagnoses and inadequate treatments.

This study focuses on noise in labels related to Myxomatous Mitral Valve Disease (MMVD), the most prevalent heart condition in small breed dogs. MMVD is evaluated through auscultation and categorized into three murmur intensity levels: \textit{mild}, \textit{moderate}, and \textit{loud/thrilling} \cite{ljungvall2014murmur}. Accurate classification is crucial for determining treatment strategies, as each category requires a different approach \cite{levine1933systolic, ljungvall2014murmur, keren2005evaluation}. Early diagnosis is essential to avoid health deterioration.

The goal of this study is to enhance AI-based classification of MMVD-related heart murmurs by refining the dataset and its labels through expert opinions. The compiled dataset consists of 140 canine heart sound recordings (HSR), annotated by five veterinary cardiologists. These recordings were used to extract features and train the classification algorithms AdaBoost, XGBoost, and Random Forest. To address label noise, the dataset was systematically cleaned and its labels were refined. The refined dataset was then used to retrain the classifiers and compare the results with those obtained for the original, noisier dataset.

This study highlights the diagnostic and therapeutic benefits of addressing noise in auscultation data labels by incorporating multiple expert opinions. Through data cleaning aimed at reducing label noise, the MMVD murmur classification performance was enhanced, especially for mild cases. The presented approach is expected to facilitate earlier and more effective interventions, ultimately improving the outcomes for dogs affected by MMVD.

\section{MATERIALS AND METHODS}
Heart sounds were collected in collaboration with Boehringer Ingelheim Vetmedica GmbH. Veterinarians from various satellite centers (SC)\footnote{A satellite center (SC) is an external facility operated by a main hospital or clinic. It provides specialized medical services closer to the patient, improving accessibility and quality of care and research.} participated in the collection. The eKuore® One Wireless Analog Digital Transducer Attachment for Standard Stethoscopes was used for this purpose.
For each dog, the veterinarians recorded $10$ seconds of data and determined whether an MMVD-related murmur was present. 
If identified, its intensity was recorded, and an echo-cardiogram was used for diagnosis, as per routine veterinary practice \cite{Bisgin2022}.

The highest murmur intensity level is characterized by the presence of chest vibrations in addition to loud audible murmurs. 
Tactile sensations cannot be accurately captured using microphone-based heart sound recordings (HSR). 
Therefore, a simplified three-class classification scheme is used for MMVD-related heart murmur intensities. 
The defined classes are: \textit{mild}, \textit{moderate}, and \textit{loud/thrilling}. 
The dataset was provided to the Fraunhofer Institute for Software and Systems Engineering 
and consisted of one HSR for each of $140$ small-breed dogs with MMVD-related heart murmurs, along with intensity assessments (\textit{SC-labels}) assigning them to one of the three categories. \\ 
\indent Initially, the provided $140$ heart sound recordings (HSR) were re-assessed. 
Each recording was evaluated up to two times by three independent leading experts in veterinary cardiology (the experts). This process resulted in a $140\times$5 label matrix. 
To ensure the reliability of the data, a novel four-step sample selection process was applied. 
This process removed samples with undesirable sound patterns as well as samples for which no clear agreement on a murmur intensity could be found. 
Following this selection, additional agreement analyses were performed. Krippendorff's $\alpha$ statistic was used for this analysis. 
The aim was to investigate potential selection biases that could occur due to multiple ratings by a single expert.
Finally, the HSR were further segmented into heart cycles. Audio features were computed from these cycles. The dataset was then divided into training and test sets. Multiple classifiers were trained and compared on these sets.

\subsection{The Labeling Mask}
\label{sec:LabelingMask}
\begin{figure}[hbt]
    \centering
	\includegraphics[width=0.9\columnwidth]{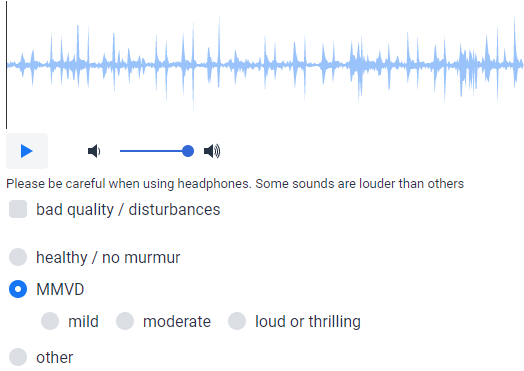}
	\caption{Labeling mask for assessing heart murmur intensity. 
 }
	\label{fig: labelingMask_fig}       
\end{figure}

Creating labeled training data is often costly and time consuming. 
When building machine learning models, data annotation tools are frequently required to efficiently collect high-quality labels. 
To systematically re-evaluate the collected $10$-second HSR, all participating experts were given access to a custom tool (\textit{labeling mask}) facilitating the assessment of the presence and intensity of MMVD-related murmurs (Fig. \ref{fig: labelingMask_fig}).
This tool allows for the asynchronous analysis of HSR, which is essential for the collection of independent labels.
By utilizing multiple independent assessments, the aim is to reduce label noise. This approach is expected to improve the performance of classification algorithms. 
It will also ensure more reliable diagnostic outcomes in cardiology practice. 
Using this tool, the cardiologists first reported whether a sample was of too \textit{bad quality} to detect heart murmurs. 
Whenever this was not the case, the experts then assessed whether it contained MMVD-related murmur (selecting \textit{healthy} or \textit{MMVD}), or if the murmur was likely due to \textit{other} causes. 
If MMVD-related murmur was detected, its intensity was ultimately classified by the experts as either \textit{mild}, \textit{moderate} or \textit{loud/thrilling}. 
Three experts conducted this process with standardized headphones for the provided $140$ HSR. 
In addition, two of the experts conducted a second pass as a basis for rater bias investigation, resulting in an HSR label matrix used for further sample selection, noise reduction, and agreement analysis.

\subsection{Sample Selection, Noise Reduction, Agreement Analysis} \label{section:data_processing}
\begin{figure}[hbt]
    \centering
    \resizebox{0.93\columnwidth}{!}{ 
    \begin{tikzpicture}[
  node distance=0.5cm and 0cm,
  every node/.style={rectangle, rounded corners, draw=black, align=center, minimum height=0.6cm, text width=0.9\columnwidth, fill=white},
  arrow/.style={thick, ->, shorten >=3pt, shorten <=3pt}
]

\node (start) [draw, fill=gray!10, minimum height=0.6cm, text width=1.05\columnwidth] {Start: \textit{Initial set of HSR (SC) provided by the Satellite Centers}};

\node (step0) [draw, fill=gray!10, minimum height=0.6cm, text width=1.05\columnwidth,below=0.4cm of start,] {Step 0: \textit{Collect expert opinions using the Labeling Mask} };

\node (step1) [below=0.4cm of step0, text width=\columnwidth] {Step 1: \textit{Remove HSR with $\geq$ 2 ``bad quality''or``other'' labels} };

\node (step2) [below=0.4cm of step1, text width=\columnwidth] {Step 2: \textit{Remove HSR with $\geq$ 2 \textit{healthy} labels}};

\node (step3) [below=0.4cm of step2, text width=\columnwidth] {Step 3: \textit{Remove HSR with 2 label classes given $\geq$ 2 times}};

\node (step4) [below=0.4cm of step3, text width=\columnwidth] {Step 4: \textcolor{black}{\textit{Remove HSR with 3 label classes given $\geq$ 1 times}}};

\node (step5) [below=0.4cm of step4, fill=gray!10, minimum height=0.6cm, text width=1.05\columnwidth] {Step 4+1: \textcolor{black}{\textit{Re-Assign HSR-labels as the most frequently mentioned murmur intensity}}};

\node (end) [below=0.4cm of step5, fill=gray!10, minimum height=0.6cm, text width=1.05\columnwidth] {End: \textit{High-quality HSR (HQ) with noise-reduced labels}};

\draw[arrow] (start) -- (step0);
\draw[arrow] (step0) -- (step1);
\draw[arrow] (step1) -- (step2);
\draw[arrow] (step2) -- (step3);
\draw[arrow] (step3) -- (step4);
\draw[arrow] (step4) -- (step5);
\draw[arrow] (step5) -- (end);

\end{tikzpicture}
}
\caption{Sample selection steps to derive high-quality HSR.}
    \label{fig: sample selection}
\end{figure}
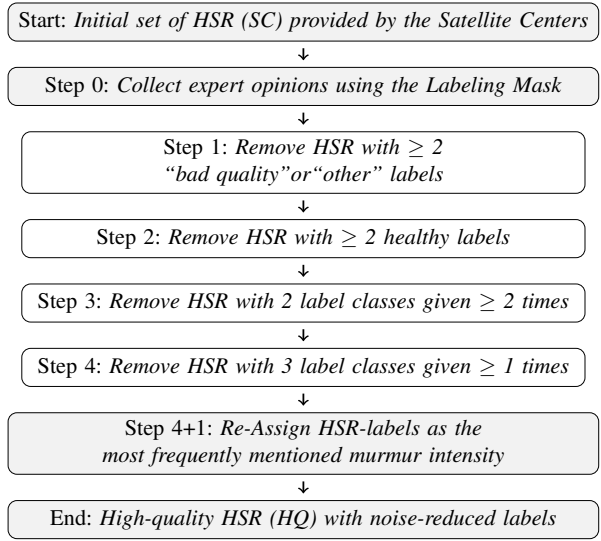

A multistep sample selection process (Fig.~\ref{fig: sample selection}) was applied to filter for those HSR for which a specific label on the presence and intensity of MMVD-related heart murmurs could be determined. This sample selection is based on the previously collected expert assessments collected with the labeling mask (Step $0$).
The four steps are as follows:
\begin{enumerate}
    \item [Step 1] \textbf{Quality Assessment}: HSR with undesirable sound patterns, either due to \textit{bad quality} or due to murmurs attributed to \textit{other} causes, were removed;
    \item [Step 2] \textbf{Murmur Detection}: HSR that were assessed to contain \textit{healthy} heart sound patterns were excluded;
    \item [Step 3] \textbf{Borderline Intensity}: If two classes are mentioned at least twice, the underlying murmurs are difficult to label, and the HSR were removed;
    \item [Step 4] \textbf{Inconclusive Intensity}: If 3 or more different assessments were made for a HSR, it was removed.
\end{enumerate}
These four selection steps are embedded in the general label-noise reduction procedure in Fig.\ref{fig: sample selection}.\\
 
Upon completion of sample selection Step $4$ (Fig.~\ref{fig: sample selection}), all remaining/selected HSR were termed as of 'high quality (HQ),' since \textit{bad quality} recordings, HSR with \textit{other} causes for audible murmurs, and HSR indicating \textit{healthy} hearts were removed. 
In consequence of these four steps, only those HSR remained for which all experts (or all but one) agreed on the presence of exactly one class of murmur intensity - either \textit{mild}, \textit{moderate}, or \textit{loud/thrilling}). 
Based on the concept of a 'majority vote,' this single diverging assessment (if existent) is overruled by the general expert consensus for a new label-option in further classification (Step 4+1). 
This new high-quality label might thereby replace the initial SC label for a given HSR, ensuring reduction of label noise.
In subsequent sections, these 'selected HSR with new labels via majority vote' will be used for classification and referred to as the high quality dataset HQ. 
Using HQ, the performance of multiple classifiers will be compared against the initial dataset of all HSR with the labels provided by the satellite centers (termed SC). 
Using these two different designs, an evaluation of the effect of sample selection on classification performance will be conducted in the following.

Given that the foundational expert assessments are subjective and manifold, better understanding the overall agreement on the subjective task of auscultating the presence and intensity of heart murmurs is crucial. In this experimental setup, multiple different experts decided on specific HSR, and some of those have done so multiple times. 
To better assess for potential rater biases throughout sample selection, and to investigate whether single expert should be asked multiple times, Krippendorff's $\alpha$ statistics \cite{krippendorffAlphaAsStandard} were calculated for the label matrix, and with each selection step:

\begin{equation}\label{eq:alphaWithWeights}
\begin{aligned}
        \alpha(X) &= \frac{p_a(X)- p_e(o_{r_1,r_2}) }{1-p_e(o_{r_1,r_2}) }
        \textrm{.}
        \nonumber
\end{aligned}
\end{equation}
Here, $X$ refers to the label matrix, $p_a$ is the observed percent agreement, whereas $p_e$ is the percent agreement expected to be observed with the difference metric $o_{r_1,r_2}$, as chance-agreement is scale-dependent. $r_1$, $r_2$ refer to two raters, since the difference metric $o_{r_1,r_2}$ is always applied pairwise.
The first statistic, $\alpha_{all}$, assesses the inter-rater agreement across all five evaluations, while $\alpha_{A}$ and $\alpha_{B}$ measure the intra-rater agreement for experts A and B. 
A value of $\alpha \approx 0$ indicates random agreement, whereas values close to ($-1$ or) $1$ suggest systematic agreement or disagreement regarding the presence and intensity of MMVD-related murmurs. 
Higher values for $\alpha_{A}$ and $\alpha_{B}$ also quantify an expert's ability to consistently assess HSR over time.

\subsection{Analysis and Classification of Heart Cycles and Murmurs} \label{sec:methods_classification} 
The overarching motivation behind Sample Selection and Noise Reduction is to increase the classification performance and robustness of the auscultation support system, facilitating more precise analyses and ultimately optimizing overall outcomes. For the sound classification under investigation, labeled audio signals are used as both training and test data.
From this point onward, the outlined classification methodology applies equally to all HSR, regardless of sample selection and label noise. 
However, distinct train-test dataset configurations will subsequently be established for all HSR and for the selected high-quality samples, together with SC labels and majority votes.
Each $10$-second recording contains $10$ to $27$ heart cycles, depending on the dog's heart rate.

By increasing the number of available samples while reducing their dimensionality, the performance and robustness of the trained classifiers can be enhanced. 
However, it is crucial to ensure that heart cycle data for a particular HSR is exclusively present in either the training or test dataset. 
Data leakage may occur if the same data is included in both datasets, leading to overly optimistic performance metrics and undermining the model's generalizability to unseen data. 
This underscores the importance of carefully separating the data to provide a realistic assessment of model performance \cite{krawczyk2016learning}. The following methodology sections summarize established approaches for preparing datasets for heart sound classification problems that utilize multiple heart cycles per patient.

\subsubsection{Preprocessing}
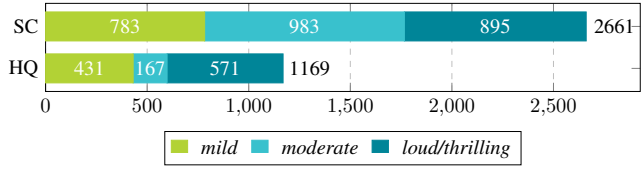
\begin{figure}[htb] 
    \centering
    \resizebox{\columnwidth}{!}{ 
    \begin{tikzpicture}
        \begin{axis}[
            xbar stacked,
            bar width=0.5cm,
            enlarge y limits=0.55, 
            ytick={2,3},
            yticklabels={HQ, SC}, %
            legend style={at={(0.5,-0.45)}, anchor=north, column sep=1ex, legend columns=6},
            xmin=0,
            xmajorgrids=true,
            grid style=dashed,
            height=3.1cm, 
            width = 1.37\columnwidth 
        ]
            
            \node[text=white] at (391.5, 3) {783};  
            \node[text=white] at (1274.5, 3) {983}; 
            \node[text=white] at (2213.5, 3) {895};  

            \node[text=white] at (215.5, 2) {431};  
            \node[text=white] at (514.5, 2) {167}; 
            \node[text=white] at (883.5, 2) {571};  
        

            \addplot[fraunhoferMint, fill=fraunhoferMint] coordinates {(431,2) (783,3) }; 
            \addplot[fraunhoferSkyBlue, fill=fraunhoferSkyBlue] coordinates {(167,2) (983,3)}; 
            \addplot[fraunhoferTurquoise, fill=fraunhoferTurquoise] coordinates {(571,2) (895,3) }; 
            
            \node at (2800,3) {2661};   
            \node at (1300,2) {1169};   

            \legend{\textit{mild}, \textit{moderate}, \textit{loud/thrilling}}
        \end{axis}
    \end{tikzpicture}
    }
    \caption{Training Heart cycles pre (SC) and post (HQ) sample selection, together with \textcolor{fraunhoferMint}{122}, \textcolor{fraunhoferSkyBlue}{43} and \textcolor{fraunhoferTurquoise}{163} test heart cycle data.}
    \label{fig:no_of_samples} 
\end{figure}

To further reduce data noise, all HSR were first filtered using a Butterworth bandpass filter with cut-off frequencies of $50$~Hz and $500$~Hz \cite{ljungvall2009use}. 
Next, the determination of heart cycles took place: Using the Hilbert function and the envelope of each individual HSR, a peak detection with dynamic parameters was applied to filter out short peaks, such as each cycle's second heart sound (S2) and noise artifacts, in order to isolate each cycle's first heart sound (S1). 
This process allowed for finding the ends of each heart cycle. Segments occurring before the first and after the last S1 sound were removed, as they correspond to incomplete heart cycles.
Through this segmentation, the $140$ HSR were transformed. 
In Fig. \ref{fig:no_of_samples}, the total number of detected heart cycles per class for both the original and adjusted datasets is shown. The number of heart cycles for the training data is described: $783$ \textit{mild}, $983$ \textit{moderate}, and $895$ \textit{loud/thrilling} heart cycle samples. Note that all heart cycles of a collected recording is assigned to the same label. 
By sample selection and label noise reduction, $70$ high-quality HSR with reliable labels were identified. In the training dataset, this corresponds to $431$ \textit{mild}, $167$ \textit{moderate}, and $571$ \textit{loud/thrilling} cycles. For the test dataset, this corresponds to the HQ dataset: $122$ \textit{mild}, $43$ \textit{moderate}, and $163$ \textit{loud/thrilling} heart cycles.\\
\indent Each extracted heart cycle was transformed into a feature vector containing time- and frequency-domain characteristics for supervised learning \cite{pantfoerder2023}. 
In the time domain, the extracted features are statistics of the amplitude series representing the respective heart cycle: Arithmetic mean, median, variance, standard deviation, skewness, kurtosis, $25$th and $75$th percentiles, maximum and minimum values, mean and median of the absolute deviation of the amplitude values, individual heart cycle duration, entropy, zero-crossing rate, tempo, pitch and crest factor.
In the frequency domain, features were derived from the discrete Fourier transform of the amplitude time series and included: dominant frequency, mean frequency, spectral entropy, bandwidth, spectral centroid, spectral roll-off, Mel Frequency Cepstral Coefficients (MFCC), Root Mean Square (RMS), pitch chroma, spectral kurtosis, spectral flux, spectral skewness, spectral flatness, spectral contrast, spectral bandwidth, spectral slope, spectral energy, spectral frequency, and spectral power. 
All these features were calculated for each heart cycle in both datasets (the original dataset and the cleaned dataset) and stored as corresponding feature vectors. 
Each heart cycle was assigned an ID, indicating its original HSR, the SC label, and the majority vote where applicable. 
Thus, one labeled feature dataset for heart murmur classification was created from all $140$ HSR and another from the selected $70$ HQ.

\subsubsection{Data Splitting and Stratification}
Given that multiple heart cycles and, therefore, feature vectors were derived from each HSR, using a standard train/test split could lead to information leakage between training and test data. 
To mitigate this risk, all extracted heart cycles were grouped by HSR. 
A test dataset was created that contained $20~\%$ of the \textit{mild}, \textit{moderate}, and \textit{loud/thrilling} classes. 
These HSR were selected such that the SC-labels matched the majority votes. 
For the corresponding heart cycles, clearly identifiable labels exist, making classification easier for all available HSR. 
Finally, the heart cycles from the remaining $126$ and $56$ HSR, along with their SC-labels and majority votes, form two stratified training datasets: SC and HQ. 

\subsubsection{Heart-Murmur Classification} 
Based on the previously described heart cycle datasets, three classification algorithms\textendash AdaBoost, Random Forest, and XGBoost\textendash were trained and evaluated. 
These algorithms were chosen for their robustness against label noise and their effectiveness in classifying feature vectors that represent physiological signals \cite{sabzevari2018vote,seiffert2014empirical,sluban2015relating,mantas2014analysis,abellan2010bagging,khoshgoftaar2010comparing}. 
In the dataset configurations SC and HQ, all three classifiers were trained with their corresponding labels, and evaluation was consistently performed using the previously selected test dataset.
The classification performance was analyzed using clinically relevant metrics such as sensitivity and specificity, calculated for each combination of classifier type, heart cycle dataset, and murmur intensity class \cite{salvi2021vascular}. 
Sensitivity measures the proportion of correctly identified positive cases, reflecting the classifier's ability to detect affected patients. 
High sensitivity is crucial for early treatment planning, as it indicates the effectiveness of the classifier in identifying affected individuals. 
Specificity, on the other hand, assesses the proportion of correctly classified negatives and indicates the classifier's ability to accurately identify patients without the condition. 
High specificity is essential to avoid false treatment planning.
In addition to sensitivity and specificity for each murmur intensity, metrics such as weighted sensitivity, weighted specificity, weighted accuracy, and the Matthews correlation coefficient (MCC) were investigated to evaluate classification performance across all three intensities (cf. Fig. \ref{fig:no_of_samples}) \cite{matthews1975comparison,chicco2023matthews}. 
These metrics enable an assessment of model performance and provide a thorough analysis of the effects of sample selection and noise reduction on the clinical applicability of machine-learning based decision support systems.

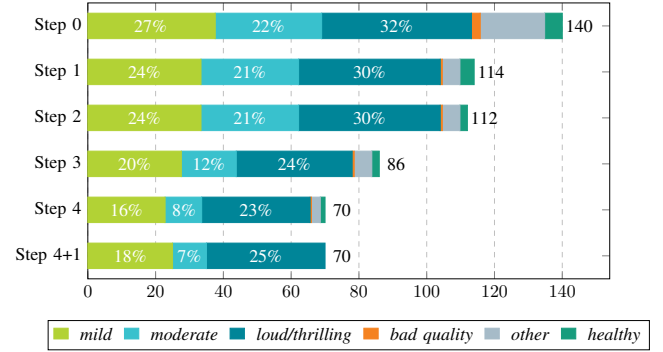
\begin{figure}[thb]
    \centering
    \resizebox{\columnwidth}{!}{ 
    \begin{tikzpicture}
        \begin{axis}[
            xbar stacked,
            bar width=0.5cm,
            enlarge y limits=0.1, 
            ytick={1,2,3,4,5,6},
            yticklabels={Step 4+1, Step 4, Step 3, Step 2, Step 1, Step 0}, 
            legend style={at={(0.5,-0.15)}, anchor=north, column sep=1ex, legend columns=6},
            xmin=0,
            xmajorgrids=true,
            grid style=dashed,
            height=7cm, 
            width = 1.37\columnwidth 
        ]


            \node[text=white] at (18.5, 6) {27\%};  
            \node[text=white] at (53.3, 6) {22\%}; 
            \node[text=white] at (91.1, 6) {32\%};  

            \node[text=white] at (16.7, 5) {24\%};  
            \node[text=white] at (47.8, 5) {21\%}; 
            \node[text=white] at (83.1, 5) {30\%};  

            \node[text=white] at (16.7, 4) {24\%};  
            \node[text=white] at (47.8, 4) {21\%}; 
            \node[text=white] at (83.1, 4) {30\%};  

            \node[text=white] at (13.8, 3) {20\%};  
            \node[text=white] at (35.7, 3) {12\%}; 
            \node[text=white] at (60.9, 3) {24\%};  

            \node[text=white] at (11.4, 2) {16\%};  
            \node[text=white] at (28.4, 2) {8\%}; 
            \node[text=white] at (49.6, 2) {23\%};  

            \node[text=white] at (12.5, 1) {18\%};  
            \node[text=white] at (30, 1) {7\%}; 
            \node[text=white] at (52.5, 1) {25\%};  

            \addplot[fraunhoferMint, fill=fraunhoferMint] coordinates {(25,1) (22.8,2) (27.6,3) (33.4,4) (33.4,5) (37.6,6)};
            \addplot[fraunhoferSkyBlue, fill=fraunhoferSkyBlue] coordinates     {(10,1) (10.8,2) (16.2,3) (28.8,4) (28.8,5) (31.4,6)}; 
            \addplot[fraunhoferTurquoise, fill=fraunhoferTurquoise] coordinates {(35,1) (32,2) (34.2,3)  (41.8,4)   (41.8,5) (44.2,6)}; 
            \addplot[fraunhoferOrange, fill=fraunhoferOrange] coordinates       {(0,1)  (0.4,2)  (0.6,3)     (0.6,4)    (0.6,5) (2.6,6)}; 
            \addplot[fraunhoferGrey, fill=fraunhoferGrey] coordinates           {(0,1)  (2.6,2)  (5.2,3)     (5.2,4)    (5.2,5) (19,6)}; 
            \addplot[fraunhoferGreen, fill=fraunhoferGreen] coordinates         {(0,1)  (1.4,2)  (2.2,3)  (2.2,4) (4.2,5) (5.2,6) };
            
            \node at (75,1) {70};   
            \node at (75,2) {70};   
            \node at (91,3) {86};   
            \node at (117,4) {112}; 
            \node at (119,5) {114}; 
            \node at (145,6) {140}; 

            \legend{\textit{mild}, \textit{moderate}, \textit{loud/thrilling}, \textit{bad quality}, \textit{other}, \textit{healthy}}
        \end{axis}
    \end{tikzpicture}
    }
    \caption{\label{fig: sample_number} Percent-wise distributions of the MMVD label in the label matrix throughout sample selection (cf. Fig.~\ref{fig: sample selection}), scaled by the count of HSR remaining at each step.}
\end{figure}

\section{RESULTS}
$140$ HSR were included in the labeling mask and evaluated by three independent cardiologists for abnormal heart sounds.

\subsection{Sample Selection} \label{sec:selection_process_results}
All $140$ HSR were re-assessed using the labeling mask to determine the presence and intensity of MMVD-related heart murmurs (Fig. \ref{fig: labelingMask_fig}). The resulting $140 \times 5$ label matrix underwent a systematic sample selection process (Fig. \ref{fig: sample selection}). 

Fig. \ref{fig: sample_number} illustrates the occurrence counts of the six selectable label classes (\textit{healthy}, \textit{mild}, \textit{moderate}, \textit{loud/thrilling}, \textit{bad quality}, and \textit{other}) before and after each selection step. The selection process led to the exclusion of $70$ samples in total. 

Specifically, the initial steps resulted in the removal of $30$ HSR due to assessments identifying them as \textit{bad quality}, containing \textit{healthy} heart sounds, or attributed to \textit{other} murmur causes. Subsequently, an additional $26$ and $14$ HSR were eliminated based on their classification into multiple label categories. 

This outcome highlights that even veterinary cardiology experts faced challenges in consistently assessing murmur intensity, as evidenced by the removal of half of the collected samples ($70/140$). 
This underscores the research challenge of preparing high-quality auscultation data for reliable murmur detection in clinical practice. 
Ultimately, only three label classes remained: \textit{mild}, \textit{moderate}, and \textit{loud/thrilling}.

\subsection{Noise Reduction}
After applying all four steps of sample selection (cf. Fig.~\ref{fig: sample selection}), the selected $70$ HSR were subjected to further refinement for noise-reduced labels. 
For each entry in the $70 \times 5$ label matrix, at least one label consistently indicated the presence of a \textit{mild}, \textit{moderate}, or \textit{loud/thrilling} MMVD-related heart murmur. 
Notably, $33$ out of the $70$ HSR exhibited complete agreement among all five labels.

Additionally, there were $14$ HSR for which a single expert assessment was recorded as \textit{healthy}, \textit{bad quality}, or \textit{other}. 
These individual assessments highlight the necessity of multiple labels per HSR; if decisions were based solely on these single assessments, $14$ HSR would have been excluded from the dataset. 
However, through majority voting, these assessments were overruled, resulting in $10$, $1$, and $3$ votes for \textit{mild}, \textit{moderate}, and \textit{loud/thrilling} intensities, respectively.

For the remaining $23$ selected HSR, noise-reduced labels were also derived via majority voting, which allowed for the overruling of single \textit{moderate} assessments with either \textit{mild} or \textit{loud/thrilling} classifications, and vice versa.

Overall, the sample selection and noise reduction processes led to a decrease in HSR classified as \textit{moderate}, which could potentially hinder effective training for this class. 
This outcome underscores the challenges associated with accurately detecting \textit{moderate} intensities solely through auscultation. 
Ultimately, a high-quality dataset of $70$ HSR was established, comprising $25$, $10$, and $35$ heart murmurs classified as \textit{mild}, \textit{moderate}, and \textit{loud/thrilling}, respectively.

\subsection{Agreement Analysis}
The resulting dataset comprises $70$ high-quality HSR with noise-reduced murmur-intensity labels. 
The study's design allowed for multiple assessments by a single expert. Potential selection biases were addressed by calculating three distinct Krippendorff's $\alpha$ statistics, $\alpha_{all}$, $\alpha_A$, and $\alpha_B$. $\alpha_{all}$ was used to quantify the baseline agreement of all expert assessments. 
Additionally, $\alpha_A$ and $\alpha_B$ measured the internal consistency of experts A and B over time. 
Fig. \ref{fig: alpha_values} presents the statistics computed for those HSR that were not excluded up to the stated selection step.
\begin{table}[htb] 
\centering
\caption{\label{fig: alpha_values} Krippendorff's $\alpha$ for inter-rater ($\alpha_{all}$) and intra-rater ($\alpha_A,\alpha_B$) reliability, on the (reduced) label matrix $X$.}
\begin{tabular}{p{0.4 \linewidth}|*{5}{>{\centering\arraybackslash}p{0.06 \linewidth}}}

    \toprule
        \textbf{Selection Steps Applied} & \textbf{None} & \textit{\textbf{1}} & \textit{\textbf{1-2}} & \textit{\textbf{1-3}} & \textit{\textbf{1-4}} \\ 
    \toprule
        \textbf{\# Heart Sound Recordings} & $140$ & $114$ & $112$ & $86$ & $70$ \\ 
    \midrule    
        \textbf{General Agreement \text{ }}\  ($\alpha_{all}$) & $0.46$ & $0.48$ & $0.47$ & $0.56$ & $0.68$ \\ 

        \textbf{Internal Agreement A} ($\alpha_A$) & $0.71$ & $0.67$ & $0.65$ & $0.71$ & $0.86$ \\  
        \textbf{Internal Agreement B} ($\alpha_B$) & $0.55$ & $0.55$ & $0.54$ & $0.58$ & $0.72$ \\ 
    \bottomrule

\end{tabular}

\end{table}

The analysis of agreement shows that for the initial dataset, the experts demonstrated a positive level of overall agreement ($\alpha_{all} = 0.46$).
Experts A and B were relatively consistent in assessing HSR concerning the six selectable label classes, with $\alpha_A = 0.71$ and $\alpha_B = 0.55$. 
An $\alpha$ value close to $0$ would indicate agreement by chance. 
The positive levels of agreement remained comparable even after excluding \textit{bad quality/other} and \textit{healthy} samples. The calculated values were: $\alpha_{all} = 0.47$, $\alpha_A = 0.65$, and $\alpha_B = 0.54$. This indicates the feasibility of having single experts re-assess HSR for heart murmurs.
To address potential selection biases favoring experts who performed more assessments, changes in $\alpha_A$ and $\alpha_B$ were analyzed. 
The results indicated that expert A rated the HSR more consistently than expert B ($\alpha_A > \alpha_B$) throughout the sample selection process. 
This suggests that there may not be a universal level of internal agreement to rely on when deciding whether to allow multiple assessments by a single expert.

However, after excluding \textit{bad quality/other} and \textit{healthy} samples, the internal agreement for both experts did not reach Krippendorff's levels of $\bar{\alpha} = 0.67$ and $\bar{\alpha} = 0.8$ for weak and strong internal agreement ($\alpha_A = 0.65$ and $\alpha_B = 0.54$). 
These findings can serve as an indicator for permitting multiple assessments by a single expert, justifying the inclusion of the $140$ HSR for this research. 
Nonetheless, the issue of selection biases warrants further discussion.

In conclusion, the systematic selection of high-quality HSR and the reduction of label noise through majority voting are crucial for developing technological auscultation support via classification algorithms for heart murmur intensity levels. 
The initial quality and labeling of collected HSR can be compromised, as evidenced by the relatively low expert agreement on the presence and intensity of clinically relevant heart murmurs over time ($\alpha_{all} = 0.46$, $\alpha_A = 0.71$, $\alpha_B = 0.55$). 
With the implemented selection and noise reduction processes, a high-quality dataset was established, demonstrating expert agreement on labels ($\alpha_{all} = 0.68$, $\alpha_A = 0.86$, $\alpha_B = 0.72$). 
These HSR and labels were subsequently used as baseline data for developing classification algorithms and clinical auscultation support systems.

\subsection{Murmur Intensity Classification and Evaluation}
\subsubsection{Selection}
Murmur intensity classification was performed using feature vectors extracted from individual heart cycles in HSR, employing the algorithms Random Forest, AdaBoost, and XGBoost. Tab. \ref{table:classifier_selection} presents the evaluation metric values achieved.
\begin{table}[htb] 
\centering
\caption{\label{table:classifier_selection}Evaluation of heart cycle classifiers based on SC labels (top; $126$ HSR) and majority votes (bottom; $54$ HSR). Metrics were computed for $328$ test heart cycles, including $122$ classified as mild, $43$ as moderate, and $163$ as loud/thrilling, with both labeling methods coinciding.}
    \begin{tabular}{l|c|c|c|c}
    \toprule
    \textbf{Classifier} & \begin{tabular}[c]{@{}c@{}} \textbf{Balanced}\\ \textbf{Sensitivity} \end{tabular} & \begin{tabular}[c]{@{}c@{}} \textbf{Balanced}\\ \textbf{Specificity} \end{tabular} & \begin{tabular}[c]{@{}c@{}} \textbf{Balanced}\\ \textbf{Accuracy} \end{tabular} & \begin{tabular}[c]{@{}c@{}} \textbf{Balanced}\\ \textbf{MCC} \end{tabular}\\
    \midrule
    \begin{tabular}[c]{@{}c@{}} {Random}\\ {Forest} \end{tabular} & 
    \begin{tabular}[c]{@{}c@{}} $38.69 \%$ \\ $71.73 \%$ \end{tabular} & 
    \begin{tabular}[c]{@{}c@{}} $75.58 \%$ \\ $91.04 \%$ \end{tabular} & 
    \begin{tabular}[c]{@{}c@{}} $65.04 \%$ \\ $90.04 \%$ \end{tabular} & 
    \begin{tabular}[c]{@{}c@{}} $0.1744$ \\ $0.6766$ \end{tabular} \\
    \midrule
    {AdaBoost} & 
    \begin{tabular}[c]{@{}c@{}} $42.72 \%$ \\ $69.74 \%$ \end{tabular} & 
    \begin{tabular}[c]{@{}c@{}} $77.54 \%$ \\ $90.04 \%$ \end{tabular}  & 
    \begin{tabular}[c]{@{}c@{}} $67.28 \%$ \\ $88.01 \%$ \end{tabular} & 
    \begin{tabular}[c]{@{}c@{}} $0.2238$ \\ $0.6166$ \end{tabular} \\
    \midrule
    {XGBoost} & 
    \begin{tabular}[c]{@{}c@{}} $42.07 \%$ \\ $80.63 \%$ \end{tabular} & 
    \begin{tabular}[c]{@{}c@{}} $75.37 \%$ \\ $93.53 \%$ \end{tabular}  & 
    \begin{tabular}[c]{@{}c@{}} $64.63 \%$ \\ $92.28 \%$ \end{tabular} & 
    \begin{tabular}[c]{@{}c@{}} $0.1883$ \\ $0.7653$ \end{tabular} \\
    \bottomrule 
    \end{tabular}\\
    \begin{tabular}{cc}
         &  
    \end{tabular}
\end{table}
The classifiers were assessed for their performance on heart cycles based on SC labels and majority votes. 
The Random Forest classifier achieved a balanced sensitivity of $38.69\%$ for SC and $71.73\%$ for majority votes, with balanced specificity values of $79.19\%$ and $88.07\%$, respectively, as well as an accuracy of $67.80\%$ and $89.93\%$. 
AdaBoost obtained balanced sensitivity values of $42.72\%$ and $69.74\%$, while its specificity increased from $77.54\%$ to $90.04\%$. 
In contrast, XGBoost exhibited the best overall performance, achieving a balanced sensitivity of $42.07\%$ and $80.63\%$, a balanced specificity of $75.37\%$ and $93.53\%$, and an accuracy of $64.63\%$ and $92.28\%$.\\
\indent Overall, the analysis indicates that XGBoost demonstrates the strongest classification performance in the majority vote scenario.

\subsubsection{Murmur Intensity Classification}
This section focuses on the classification of murmur intensities, specifically \textit{mild}, \textit{moderate}, and \textit{loud/thrilling}. 
The classification was performed using feature vectors extracted from individual heart cycles, employing algorithms such as Random Forest, AdaBoost, and XGBoost.

Tab. \ref{table-random-forest} presents the evaluation metrics for the XGBoost classifier based on SC labels and majority votes. 
The metrics were computed for a total of 328 test heart cycles, consisting of $122$ classified as \textit{mild}, $43$ as \textit{moderate}, and $163$ as \textit{loud/thrilling}, where both labeling methods coincided. 

For the \textit{mild} class, the sensitivity was $37.71\%$ for SC labels and improved to $90.98\%$ for majority votes. 
The specificity was $76.70\%$ for SC and increased to $93.69\%$ for majority votes. The accuracy for the \textit{mild} class was $62.20\%$ for SC labels and rose to $93.69\%$ when using majority votes. 
The Matthews correlation coefficient (MCC) improved from $0.1540$ to $0.8440$.

For the \textit{moderate} class, sensitivity was $30.23\%$ for SC labels and increased to $55.81\%$ for majority votes. 
The specificity was $64.56\%$ for SC and rose to $97.19\%$ for majority votes. The accuracy was $60.06\%$ for SC labels and increased to $91.77\%$ for majority votes, while the MCC changed from $-0.037$ to $0.6029$.

In the case of the \textit{loud/thrilling} class, sensitivity was $58.28\%$ for SC labels and improved to $95.09\%$ for majority votes. 
The specificity was $84.84\%$ for SC and $89.69\%$ for majority votes. The accuracy for the \textit{loud/thrilling} class was $71.65\%$ for SC labels and increased to $92.38\%$ for majority votes. 
The MCC improved from $0.4477$ to $0.8489$.

Overall, the analysis indicates that XGBoost demonstrates the strongest classification performance in the majority vote scenario.

\begin{table}[htb]
    \centering
    \caption{XGBoost evaluation metrics for SC labels (top; $126$ HSR) and majority votes (bottom; $54$ HSR) were computed for $328$ test heart cycles: $122$ \textit{mild}, $43$ \textit{moderate}, and $163$ \textit{loud/thrilling}, where both labels matched.}
    \label{table-random-forest}
    \begin{tabular}{l|c|c|c|c}
    \toprule
    \textbf{Class}
    \textbf{} & \textbf{Sensitivity} & \textbf{Specificity} & \textbf{Accuracy}  & \textbf{MCC} \\
    \midrule
    \textit{mild} & 
    \begin{tabular}[c]{@{}c@{}} $37.71\%$ \\ $90.98\%$ \end{tabular} & 
    \begin{tabular}[c]{@{}c@{}} $76.70\%$ \\ $93.69\%$ \end{tabular}  & 
    \begin{tabular}[c]{@{}c@{}} $62.20\%$ \\ $93.69\%$ \end{tabular} & 
    \begin{tabular}[c]{@{}c@{}} $0.1540$ \\ $0.8440$ \end{tabular} \\
    \midrule
     \textit{moderate} & 
    \begin{tabular}[c]{@{}c@{}} $30.23\%$ \\ $55.81\%$ \end{tabular} & 
    \begin{tabular}[c]{@{}c@{}} $64.56\%$ \\ $97.19\%$ \end{tabular}  & 
    \begin{tabular}[c]{@{}c@{}} $60.06\%$ \\ $91.77\%$ \end{tabular} & 
    \begin{tabular}[c]{@{}c@{}} $-0.037$ \\ $0.6029$ \end{tabular} \\
    \midrule
    \begin{tabular}[c]{@{}c@{}} \textit{loud/}\\ \textit{thrilling} \end{tabular} & 
    \begin{tabular}[c]{@{}c@{}} $58.28\%$ \\ $95.09\%$ \end{tabular} & 
    \begin{tabular}[c]{@{}c@{}} $84.84\%$ \\ $89.69\%$ \end{tabular}  & 
    \begin{tabular}[c]{@{}c@{}} $71.65\%$ \\ $92.38\%$ \end{tabular} & 
    \begin{tabular}[c]{@{}c@{}} $0.4477$ \\ $0.8489$ \end{tabular} \\
    \bottomrule
    \end{tabular}
\end{table}

\section{DISCUSSION}
Accurate classification of sound intensity is crucial for timely analyses. 
While the XGBoost algorithm shows high sensitivity for \textit{loud/thrilling} cases, its lower sensitivity for \textit{moderate} sounds is concerning, as clinical decision-making based on this assessment is not yet reliably possible. 
This may result from the tendency of analysts to classify \textit{moderate} cases as \textit{mild} or \textit{loud/thrilling}. 
The filtering of samples during selection also contributed to class imbalance, based on the data in Fig.\ref{fig: sample selection}. 
This might well be due to subjective nature of what defines \textit{moderate} murmurs for practicing clinicians. 

This research addressed the core challenge of label quality in machine learning for clinical practice. 
Future studies could utilize strategies such as data augmentation. They could also focus on engineering additional audio features for the derived dataset. 
Collecting more high-quality data might help mitigate the data imbalance after sample selection. 
Alternative machine learning models, such as Convolutional Neural Networks (CNN) on spectral images or Recurrent Neural Networks (RNN) on the audio time series data, might also improve classification performance. Instead of manually selecting features by experts, deep learning methods could be used to automatically identify relevant features in the data. 

These approaches could provide novel insights into the characteristics of the murmur patterns and especially hidden patterns. This could further improve clinical decision support for patients and users, and thus make decision-support more reliable.

Selection biases pose a challenge for training and evaluating machine learning models, particularly in applications where data quality is critical \cite{kohavi1995cross}. 
In this study, biases may have arisen from the annotation and filtering processes. The awareness of evaluators might have influenced their assessments. 
Moreover, the filtering process (see Fig. \ref{fig: sample selection}) may have exacerbated these biases, as valuable samples labeled as \textit{healthy} or with conflicting annotations were removed. 
To mitigate these biases, a diverse and representative sampling of sound intensities is essential. Future studies should employ stratified sampling techniques to ensure balanced representation across categories.

The applicability of the developed algorithms for classifying sounds is of great importance, as accurate classifications impact overall outcomes. 
The promising performance of the XGBoost algorithm in classifying sounds by intensity highlights its potential as a valuable tool. 
Enhancing early detection of sound variations, particularly in mild and moderate cases often overlooked, is crucial for facilitating timely interventions.
Overall, the results indicate that relabeling data through independent observers can improve prediction accuracy and address subjective perceptions of intensity. 
The model effectively optimized the detection of the three noise intensities, reinforcing the effectiveness of the XGBoost algorithm in classifying sounds \cite{zadrozny2002transforming}.

The sensitivity improvements for \textit{mild} and \textit{loud/thrilling} cases suggest that the model is well-suited for applications where timely analyses are vital. 
However, the decreased sensitivity for \textit{moderate} sounds indicates a need for further research. 
Future work should focus on optimizing classifier parameters and incorporating additional features that capture subtle variations in sounds. However, it is important to note that this paper did not include a cross-validation step due to limited data availability. The number of heart sound recordings was insufficient to create a separate validation dataset without significantly reducing the training data. Moreover, the study prioritized initial analysis and methodology development, with validation planned for future research. Thus, this work can be considered a pilot study aimed at gaining fundamental insights before implementing comprehensive validation measures.

\section{CONCLUSION} This study introduced a systematic approach to reduce noise in subjective medical labels using multiple expert opinions gathered with a digital labeling tool. 
The approach's effectiveness was demonstrated for the use case of MMVD murmur classification based on auscultation data.

The XGBoost algorithm's high detection performance for \textit{mild} and \textit{loud/thrilling} murmurs supports the effectiveness of sample selection and noise reduction. However, the lower sensitivity for \textit{moderate} murmurs indicates the need for further refinement. Future efforts should focus on incorporating features that capture subtle heart sound variations, applying different classifier types such as CNNs, and optimizing hyperparameters. Continued labeling by veterinary cardiologists will further expand the dataset and enhance the classification capabilities.

The presented label noise reduction strategy is expected to enhance the quality of AI-based diagnostics in veterinary medicine and sets a precedent for similar research in other medical fields. An additional application may exist due to the comparability of canine and human heart sounds as well as the acoustical similarity between canine MMVD and human mitral valve prolapse. These factors may open the possibility of employing fundamental aspects of the developed algorithms for both dogs and humans. Furthermore, transfer learning approaches may allow the reuse of training results across both areas.


\section*{ACKNOWLEDGMENT}
This project was funded by Boehringer Ingelheim Vetmedica GmbH.
\vspace{1em}
\\
\noindent\textsf{\textit{Author Statement}}
\\
The authors declare that no human ethics approval was required for this study. The approval by the Institutional Animal Care and Use Committee or other approval declaration was granted by Ludwig Maximilian University Munich and the Swedish University of Agricultural Sciences, Uppsala, Sweden, for all centers.

\bibliographystyle{IEEEtran}
\bibliography{bibliography}
\end{document}